\documentclass[runningheads]{llncs}
\usepackage[T1]{fontenc}
\usepackage{graphicx}
\usepackage{hyperref}
\usepackage{color}
\usepackage{cite}

\hypersetup{
	colorlinks=true,
	linkcolor=blue,
	anchorcolor=black,
	citecolor=blue,
}

\usepackage{xcolor}
\usepackage{float}
\usepackage{marvosym}
\usepackage{booktabs}
\usepackage{multirow}
\usepackage{orcidlink}
\usepackage{xspace}
\usepackage{subfigure}
\usepackage{booktabs}
\usepackage{amssymb}
\usepackage{amsmath}
\usepackage{float}
\usepackage{amsfonts}
\usepackage{threeparttable}

\newcommand{\fullref}[2]{\hyperref[#2]{#1~\ref*{#2}}}
\newcommand{\mypara}[1]{{\vspace{0.1mm}\noindent\textbf{#1}}}

\begin{document}
\title{RobotDiffuse: Diffusion-Based Motion Planning for Redundant Manipulators with the ROP Obstacle Avoidance Dataset}
\titlerunning{RobotDiffuse: Diffusion-Based Motion Planning}
% \title{\LARGE \bf
% RobotDiffuse: Motion Planning for Redundant Manipulator based on Diffusion Model using Large-Scale Obstacle Avoidance Dataset
% }

% \author{Wentao Song\inst{1}\orcidID{0000-1111-2222-3333} \and
% Ye Tao\inst{2,3}\orcidlink{1111-2222-3333-4444}\textsuperscript{(\Letter)}\and
% Rui Wang\inst{3},$^{\orcidlink{0009-0004-5900-605X}}$}
% \author{Anonymous Author(s)}
% \authorrunning{Anonymous Author(s)}
% \institute{Paper ID: [0044]}

\author{Xudong Mou\inst{1}\textsuperscript{,\orcidlink{0009-0005-1445-3742}} \and
Xiaohan Zhang\inst{1}\textsuperscript{,\orcidlink{0000-0002-9575-9604}} \and
Tiejun Wang\inst{1}\textsuperscript{,\orcidlink{0000-0003-0391-8661}} \and
Tianyu Wo\inst{2,3}\textsuperscript{,}\thanks{The corresponding author is Tianyu Wo.}\textsuperscript{,\orcidlink{0000-0002-5331-3364}}\and
Cangbai Xu\inst{1}\textsuperscript{,\orcidlink{0009-0004-1361-0753}} \and
Ningbo Gu\inst{3}\textsuperscript{,\orcidlink{0009-0001-1220-0099}} \and
Rui Wang\inst{1}\textsuperscript{,\orcidlink{0009-0004-5900-605X}} \and
Xudong Liu\inst{1,3}\textsuperscript{,\orcidlink{0000-0001-8566-660X}}
}

\authorrunning{M. Xu et al.}


\institute{School of Computer Science and Engineering, Beihang University, Beijing, China \and
 School of Software, Beihang University, Beijing, China \and
 Hangzhou Innovation Institute, Beihang University, Hangzhou, China
\\
\email{\{mxd,zhangxh,wtj,woty,zy2306113,guningbo,ruiking,liuxd\}@buaa.edu.cn}
}
\maketitle              % typeset the header of the contribution

\begin{abstract}
Redundant manipulators, with their higher Degrees of Freedom (DoFs), offer enhanced kinematic performance and versatility, making them suitable for applications like manufacturing, surgical robotics, and human-robot collaboration. However, motion planning for these manipulators is challenging due to increased DoFs and complex, dynamic environments. While traditional motion planning algorithms struggle with high-dimensional spaces, deep learning-based methods often face instability and inefficiency in complex tasks.
This paper introduces RobotDiffuse, a diffusion model-based approach for motion planning in redundant manipulators. By integrating physical constraints with a point cloud encoder and replacing the U-Net structure with an encoder-only transformer, RobotDiffuse improves the model's ability to capture temporal dependencies and generate smoother, more coherent motion plans. 
We validate the approach using a complex simulator and release a new dataset, Robot-obtalcles-panda (ROP), with 35M robot poses and 0.14M obstacle avoidance scenarios. The highest overall score obtained in the experiment demonstrates the effectiveness of RobotDiffuse and the promise of diffusion models for motion planning tasks.
The dataset can be accessed at \url{https://github.com/ACRoboT-buaa/RobotDiffuse}.

\end{abstract}

\section{Introduction}
Redundant manipulators have garnered significant attention in academia and industry because of their ability to enhance kinematic performance, agility, and dynamic capabilities with higher Degrees of Freedom (DoFs). 
They can be deployed in various applications \cite{mu2022hyper}, including advanced manufacturing, precision assembly, surgical robotics, autonomous material handling, and human-robot collaboration. 
Motion planning is crucial for optimizing the robot’s tasks in real-world settings, ensuring stability and smooth operation while considering environmental constraints. 
Higher DoFs introduce challenges in inverse kinematics during motion planning. Additionally, the complexity and variability of obstacles and the diverse starting and ending points of tasks extend beyond a single, fixed environment.

Motion planning algorithms focus on building a path from the initial to the target position, avoiding environmental obstacles, and satisfying potential constraints while optimizing the path according to certain criteria~\cite{mukadam2018continuous}. Planning algorithms fall mainly into search-based, sampling-based, and intelligence-based. Search-based algorithms can find the theoretical optimal solution but suffer from node explosion in high-dimensional spaces. Sampling-based algorithms use probabilistic completeness to address high-dimensional spaces and have improved sampling strategies to reduce search space and enhance solution efficiency. However, they are still limited by high dimensionality and redundancy, requiring significant computation time for complex problems.

\begin{figure*}[t]
    \centering
    \subfigure[Sample-based motion planning - RRT*]{
    \begin{minipage}[t]{\linewidth}
    \centering 
    \includegraphics[width=\linewidth]{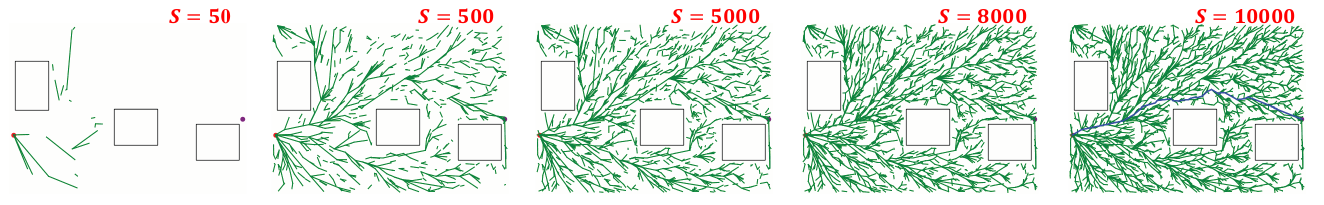}
    \label{motivation_1}
    \end{minipage}
    }%
    \\
    \subfigure[Diffusion-based motion planning - RobotDiffuse]{
    \begin{minipage}[t]{\linewidth}
    \centering
    \includegraphics[width=\linewidth]{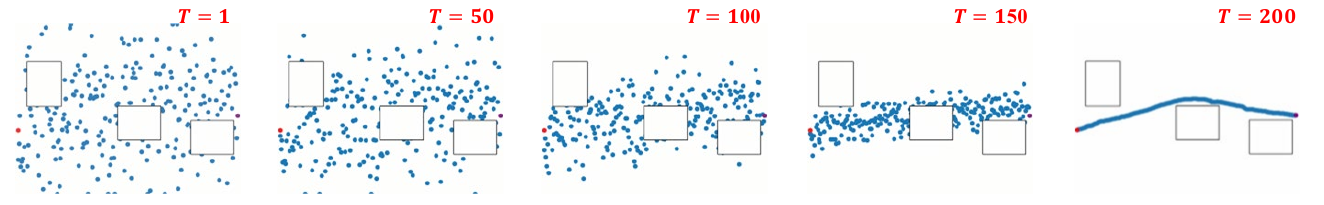}
    \label{motivation_2}
    \end{minipage}
    }%
    \centering
    % \vspace{-0.3em}
    \caption{Schematic of the motivation in this paper. Comparing the sample-based and our diffusion-based approaches. $S$ and $T$ denote the number of sampling and diffusion steps, respectively. Sample-based methods require more time to generate a path and often involve frequent backtracking. }
    \vspace{-0.5cm}
    \label{motivation}
\end{figure*}
\vspace{-0.3em}

Deep learning-based methods learn from experience and make quick decisions by leveraging problem similarities and prior knowledge. Mainstream deep learning motion planning approaches use neural networks to guide sampling directions, iteratively generating the next move until the robot reaches the target. While this simplifies motion planning, as movement is only determined by the target and environment, it has drawbacks: the solving time fluctuates with network modeling capacity and task complexity, and it overlooks previous directions while sampling points, leading to instability and unnecessary wear or inefficiency. Modeling motion planning as a motion generation task, where the neural network generates global movement in one step, offers an effective solution.
 
The Diffusion Model (DM) is a generative model derived from non-equilibrium thermodynamics~\cite{sohl2015deep}. 
In recent years, it has achieved excellent results in video, image, and speech generation, surpassing the Generative Adversarial Network (GAN). The core of the diffusion model is to simulate the evolution of data distribution. During training, it repeatedly adds noise and denoises to learn the transformation process from the Gaussian distribution to a complex distribution. 
As shown in \fullref{Fig.}{motivation}, the sample-based methods require extensive exploration of the solution space and backtracking~\cite{karaman2011sampling}, whereas diffusion models proposed in this paper can quickly generate feasible solutions. The generated results have the advantages of strong generalization and high diversity. 
However, deploying DM on motion planning faces the following challenges: 1) The U-Net in DM excels at extracting local features but is weak in capturing temporal dependencies. 2) DM ignores physical world constraints and the unique motion characteristics of each robot. Recent attempts~\cite{carvalho2023motion, saha2024edmp} employed physical guidance during DM sampling but have not directly addressed either of these challenges, which will be described in related work. 
This paper proposes a DM-based motion planning approach for redundant manipulators named RobotDiffuse.
It describes the physical constraints during training with a well-trained point cloud encoder, accurately capturing obstacles regardless of rotations while filtering out redundant information. Then, it uses an encoder-only network to replace the U-Net structure in the DM to learn more temporal dependencies for longer motion planning. 
We deploy a PyBullet-based simulator~\cite{panerati2021learning} and generate a large and complex dataset, Robot-obtalcles-panda (ROP), proposed in this paper. 
Experimental results demonstrate the effectiveness of our approach and the promise of diffusion models in motion planning tasks.
The contribution is listed below.

\begin{itemize}
    \item We propose a motion planning approach for redundant manipulators, RobotDiffuse, which generates consistent moving directions compared to sample-based methods and is the first to introduce constraints during training among DM-based methods, as far as we know. 
    \item We leverage an encoder-only Transformer architecture to fully capture the temporal dependencies in the movement sequences. This enhances the coherence of the output planning and significantly improves the model's ability to understand and navigate complex scenes.
    \item We release a highly complex motion planning dataset ROP, tailored for non-desktop scenarios, comprising 1.6 million frames of poses and 200 diverse obstacle-avoidance scenarios. Additionally, we provide the encoding methodology used for its generation.
\end{itemize}

\section{Related Work}
This section reviews robot motion planning research across search, sampling, and intelligence-based paradigms, with a focus on recent diffusion-based approaches. 
Classical Planning Paradigms Search-based algorithms, such as $A^*$~\cite{hart1968formal} and $D^*$~\cite{stentz1994optimal}, achieve global optimality by combining heuristic cost functions with actual distances. 
However, they suffer from the curse of dimensionality, where performance degrades sharply as search nodes increase in high-dimensional spaces. Sampling-based algorithms, including PRM and RRT, address high-dimensional challenges by connecting collision-free states in the configuration space. 
While $RRT^*$~\cite{karaman2011sampling} and its variants (e.g., Informed $RRT^*$, $RRT$-Connect) introduce asymptotic optimality through path rewiring, they often struggle with narrow passages and the balance between path quality and computational efficiency in time-sensitive scenarios. 
Intelligence-based \& Diffusion Models Learning-based methods like MPNet~\cite{qureshi2020motion} utilize neural networks for multi-axis planning but frequently encounter cumulative prediction errors. 
Recently, diffusion models have emerged as a powerful alternative. While existing frameworks, such as MPD~\cite{carvalho2023motion} and EDMP~\cite{saha2024edmp}, incorporate task-specific priors or cost functions during inference, many still rely on traditional U-Net architectures and lack physical constraints during the training phase. 
This often leads to suboptimal performance in complex, physically-constrained 3D environments.

\section{Preliminaries}

\subsection{Problem definition}
\textbf{Notations.}  The posture state of the robotic arm $x \in \mathbb{R}^D$, $D$ is determined by its DOF. A trajectory, $X$, is a sequence of joint states over $N$ frame, i.e. $X\in\mathbb{R}^{D\times N}$. We use $X_t $ to represent the noise $T$ sequences of a series of robot joints under $X_1, X_2,... X_T$. The robot configuration space is defined as $CS\subset \mathbb{R}^D$ by the obstacle space $CS_{obs}$ and the accessible space $CS_{free}=CS\backslash CS_{obs}$. The robot's workspace is denoted as $WS\subset\mathbb{R}^M$, where $M$ is the workspace dimension. Similar to configuration space, workspaces can also be divided into obstacle Spaces $WS_{obs}$ and accessibility Spaces $WS_{free}=WS\backslash{WS}_{obs}$. For a motion task, the robot moves in any state $x\in CS_{free}$. The obstacle point cloud is represented as $I \in \mathbb{R} ^ {K \times 3} $, where $K$ is the number of points in the point cloud. We are interested in solving an optimization problem in the collision-free space $X\subset C_{free}$ after a given task,

\begin{equation}
    \mathop{min}\limits_{X} J(X;o,CS),
\end{equation}
where the vector $o$ represents the hyperparameter of the cost function $J$.

\begin{figure*}[th]
    \centering
    \includegraphics[width=\linewidth]{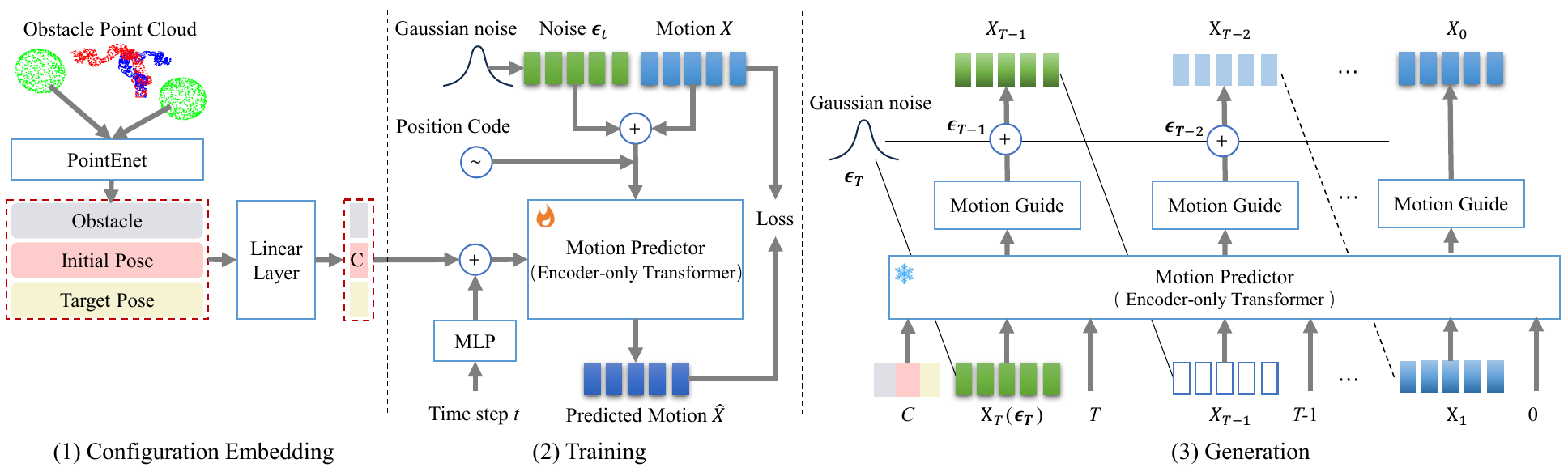}
    \caption{RobotDiffuse overview. It contains three steps: embedding, training, and generation. (1) The obstacle point net is embedded into the latent space as $z$ by the PointEnet component.
    Together with the init pose $CS_{init}$, the target pose $CS_{goal}$ comprises a condition code $C$. (2) $C$ is randomly masked for classifier-free learning and then projected with $t$ into the input token $z_{tk}$. Given a noise and motion $X$, the motion predictor based on an encoder-only transformer predicts the original clean motion $\hat{X}^{1:N}$. (3) Given a condition $C$, we sample random noise $X_T$ at the dimensions of the desired motion, then iterate from $T$ to $1$. At each step $t$, the predictor predicts a clean sample $\hat{X}$ and then adds noise, bringing it back to $X_{t-1}$.
    }
    \label{framework}
    \vspace{-2em}
    \end{figure*}

\section{Method}
The framework of our proposed robotDiffuse is shown in \fullref{Fig.}{framework}. For the task of robot motion generation, the dataset comprises pairs of $(X, C)$, where $C$ is the description of the motion sequence $X$, including obstacle point cloud environment $I\in {WS}_{obs}$ and the robot's initial and final pose. 
% Obstacle point cloud environment $I\in {WS}_{obs}$. 
The insight lies in the generation of high-quality and diverse robot motion sequences with a desired conditional input $C$, which contains three steps: embedding, training, and sampling. 
First, the task conditions are embedded in the latent space based on the pre-trained point-cloud encoder PointEnet, allowing the model to understand the environmental and task characteristics.
Then, Gaussian noises are added to the latent representation, and the denoising process is performed with a transformer network. The transformer encodes the environment embedding $Z$, initial pose $CS_{init}$, and target pose $CS_{goal}$ with the parameter $\theta ^p$ to generate a motion trajectory $X$. Finally, the reconstructed latent representations are restored to trajectories and are optimized with our guidance-based approach.

% \subsection{Task Semanticization based on point cloud}
\subsection{Condition embedding}
To eliminate the interference of irrelevant information, such as texture during obstacle avoidance motion planning, and inspired by \cite{liu2024gram}, we use a Compression AutoEncoder (CAE) as the point cloud encoder for self-training.
This architecture can learn the robust feature space required for planning and control and perform better than other available coding techniques. 
CAE uses common reconstruction losses and regularization of encoder parameters. To improve the generalization ability, this part of the parameter will be frozen after training and will not be optimized in subsequent training.
\begin{equation}
    l_{AE}(\theta^e,\theta^d)=\frac{1}{K}\sum_{i\in I}\Vert i-\hat{i} \Vert^2+\lambda{\sum_{ij}}(\theta_{ij}^2)^2,
\end{equation}
% $$l_{AE}(\theta^e,\theta^d)=\frac{1}{N_{obs}}\sum_{x\in D_{obs}}\Vert x-\hat{x} \Vert^2+\lambda{\sum_{ij}}(\theta_{ij}^2)^2$$  
%其中 $\theta^e$,$\theta^d$是编码器和解码器的参数，$\lambda$是正则化系数，编码$\hat{x}$表示重构点云。训练数据集下的障碍物点云空间$x\subset X_{obs}$。
where $\theta^e$, $\theta^d$ are the parameters of the encoder and decoder, $\lambda$ is the regularization coefficient, and the encoding $\hat{x}$ represents the reconstructed point cloud. PointEnet with 3-layer Perceptrons uses the parameter $\theta ^e$to compress the obstacle environment information $I\subset WS_{obs}$ into the potential space $Z$, that is,
\begin{equation}
   Z \leftarrow PointEnet(I;\theta^e).
\end{equation}

\subsection{ Diffusion Model for Motion Generation}

Generative Adversarial Networks (GANs) are known for generating high-quality data through adversarial training, which can be effective when working with limited data. However, they often struggle to generalize under complex conditions. Moreover, predicting robot motion sequences based on tasks is not a simple one-to-one mapping, as multiple motion sequences can correspond to a given task. Inspired by the recent success of diffusion models in image generation, we apply a diffusion model to generate task-conditioned robot joint sequences. This probabilistic framework promotes diverse motion generation, and we leverage the denoising diffusion probabilistic model to facilitate this motion generation process.

RobotDiffuse contains an encoder-only transformer instead of the common Unet structure of diffusion models~\cite{ho2020denoising, ramesh2022hierarchical} since it captures more temporal dependencies of motion sequences of any length, as prior work in human motion planning proved~\cite{tevet2023human}.
The transformer takes 2 inputs, including time step $t$ and condition $C$. Then, it predicts a joint sequence from $X_1,... ,X_T$. We project the timestep $t$ and condition $C$ into the transformer dimension via a separate feedforward network and integrate them into the token $z_{tk}$. The joint sequence with noise $X_t$ is also linearly projected to the transformer size and is embedded and summed with a standard position coding. Input the projected noise to the decoder by entering $z_{tk}$ across the attention layer, project the result back to the original motion dimension, and as a prediction $\hat{X}_0$.
Given a task $C$, the pre-trained point cloud Enet encodes $C$ as a potential space $Z$. Now,t he forward diffusion process can be expressed as follows:
\begin{equation}
z_t=\sqrt{\overline\alpha_t}z_0+\sqrt{1-\overline\alpha_t}\epsilon_t,\epsilon_t\sim\mathcal{N}(0,\textbf{I}).
\end{equation}
$z_T$ and the input $X$ are fed into the diffusion model. After the reverse denoising process of step T, the final variable $\hat{z}$ is reconstructed into the robot motion path $\hat{X}$.

We optimize the diffusion model through a two-stage optimization process combined with motion feedback.

\mypara{Optimize via Robot Feedback}
Learning the mean can be reparameterized as learning to predict the original data $x_0$. Instead of predicting $\epsilon_t$ as formulated by \cite{ho2020denoising}, we follow \cite{ramesh2022hierarchical} and predict the signal itself. 
The training loss is defined as a reconstruction loss of $x_0$:
\begin{equation}
    \mathcal L_{simple}=\mathbb{E}_{x_0,n}\Vert \hat{x}_\theta(x_n,n,c)-x_0 \Vert_1.
\end{equation}
As shown in  \fullref{Fig.}{framework}, in denoising step $n$, we concatenate joint pose condition $c$ with joint pose representation $X_1^n,..., X_T^n$ at noise level $n$, combined with noise embedding as input to a diffusion model, and estimate $x_0$.

%RobotDiffuse使用复合损失函数进行训练，包含3个组成部分：构型空间损失和几何任务空间损失（用于强制准确预测）和碰撞损失（用于防止灾难性行为）
RobotDiffuse is trained using a composite loss function with 3 components: configuration space loss, geometric task space loss (to force accurate predictions), and collision loss (to prevent catastrophic behavior).
\begin{equation}
\begin{aligned}
\mathcal L &=\lambda_{joint}*\mathcal L_{simple} +\lambda_{point}*\mathcal L_{point} +\lambda_{collision}*\mathcal L_{collision}.
\end{aligned}
\end{equation}

$\mathcal L_{simple}$ represents the L1 loss of joint rotation, and $\mathcal L_{point}$ represents the L1 loss of the robot point cloud, where FK is the Forward Kinematics, which can calculate the position and attitude of any link of the robot given the joint rotation of the robot.

\begin{equation}
\begin{gathered}
\mathcal L_{joint}=
\mathcal L_{simple}=\frac{1}{N} \sum_{i=1}^N\left\|\hat{x}_0^i-x_0^i\right\|^2,
\end{gathered}
\end{equation}

\begin{equation}
   \mathcal L_{point }=\frac{1}{N} \sum_{i=1}^N\left\|F K\left(\hat{x}_0^i\right)-F K\left(x_0^i\right)\right\|^2,
\end{equation}

$\mathcal L_{collision}$ represents the collision loss of the point cloud,
\begin{equation}
    \mathcal L_{collision }=
\frac{1}{N} \sum_{i=1}^N \operatorname{h}\left(FK(\hat{x}_0^i),I\right).
\end{equation}

The loss of configuration space has been extensively used in previous work~\cite{fishman2023motion}. Geometric task space loss can effectively capture the cumulative error of the motion chain. We construct a distance function $D$. During training, for a given closed surface, $D$ computes the minimum distance between the robot's point cloud and the surface and returns a negative value if the point cloud is inside the closed surface
\begin{equation}
    h(ws,I)= \begin{cases}S-D(ws,I) \quad if 
 D(ws,I)\leq{S}\\0 
 \quad if D(ws,I) > S\end{cases},
\end{equation}
 % $S$ is the safe distance and $ws\in WS$ is a point representing the robot link, consisting of its position information in terms of x, y, and z.
 where $S$ represents the safe distance, and $ws \in WS$ denotes a point corresponding to the robot link, including its position information in the 3-dimensional coordinates.

% \mypara{Motion Guided Generation}
\subsection{Motion Guided Generation}
In addition, the motion sequence generated by DDPM is presented in its explicit form rather than being compressed into the latent space. Given a task, when sampling the model, the noise term $\epsilon_{\theta}(x_n,n, condition)$ determines the de-noising direction of the motion sequence in each iteration. 
Therefore, we apply physical constraints on the denoising process to guide the generation process and promote the trajectory to maintain kinematic validity. We combine several different cost functions to guide diffusion. Specifically, at each step of the denoising time, we modify the intermediate trajectory $\tau_t$ of the denoiser prediction under the first $t$ diffusion time step. 
A ``noise interpolation'' method is proposed to control the initial and final attitude of the robot motion sequence; that is, the prefix frame and suffix frame of $\hat{x}_0$ are covered by the initial and final attitude of the task. This encourages motion generation to be consistent with the original input, solves the motion intermediate problem~\cite{harvey2020robust}, and causes the network to produce a smooth trajectory in the start-goal pair.

\section{Experiments}

\subsection{Datasets}

To verify our approach in more complex obstacle avoidance scenarios, we specifically made a dataset pipeline and produced a dataset ROP using PyBullet. 
ROP consists of more than 35 million frames of robot poses and more than 0.14 million obstacle avoidance scenarios. Each scene sets challenging spheres and cube obstacles in random positions. This is a highly challenging setup compared to fixed obstacles in most current work. 

This paper proposes the Shared-Tree Informed RRT* algorithm for generating appropriate paths for the dataset, introducing node security defense and shared spanning tree strategies based on Informed RRT*. These innovations improve the performance and quality of traditional motion planning algorithms when applied to redundant manipulators. Each motion planning problem consists of three components: the obstacle scene, initial pose $CS_{init}$, and target end position $WS_{goal}$. Using inverse kinematics, a set of target poses $CS_{goals}$ can be generated. A motion planning algorithm then computes a collision-free joint sequence from the initial pose to the target pose. The shared spanning tree strategy enables a simultaneous search of ${CS_{goals}}$, allowing for reuse of the generated tree and ultimately providing an optimal path for multiple goals, thereby further enhancing both generation efficiency and path quality. The node security defense strategy addresses the limitations of traditional algorithms, which often generate paths too close to obstacles, by ensuring the robot avoids collisions and maintains a safe distance from obstacles during movement.
The dataset and visualizations are available at \url{https://github.com/ACRoboT-buaa/RobotDiffuse}.

\subsection{Baselines}
The proposed method is compared with several baselines, including the classical sample-based RRT*, intelligence-based approaches, and diffusion-based methods.  
RRT* \cite{karaman2011sampling} is a random sample-based approach, widely recognized in path planning and known for its ability to provide asymptotically optimal paths in high-dimensional spaces.
MPNet \cite{qureshi2020motion} uses autoencoders and multilayer perceptrons for intelligent sampling-based motion planning.
MPINet \cite{fishman2023motion} uses a PointNet++ point cloud deep network to model tasks with obstacle point clouds and robot endpoint clouds, and utilizes a multi-layer perceptron for intelligent motion planning.
MPD \cite{carvalho2023motion} is DM-based, pretrains an unconditional motion generator in a fixed multi-obstacle scenario and guides the generation to meet the task requirements through a cost function during sampling.

\subsection{Evaluation Metrics}

In addition to the commonly used success rate, we consider the collision rate and length of the path to evaluate the manipulator's trajectory in more complete views. 
\textbf{Success rate.} A trajectory is considered successful if it avoids physical violations and exhibits stable motion, with the final task space posture deviating by no more than 1 cm and 15° from the target. Physical violations include collisions, exceeding joint limits, and erratic behavior (such as excessive twitching).
\textbf{Collision rate.} The rate of fatal collisions, which involves both the manipulator itself and surrounding scene obstacles.
\textbf{Path length.} For redundant robots, the total motion path is the sum of the individual link trajectories.
\textbf{ComScore.} Considering that the generation time should be comprehensively taken into account, we have designed a comprehensive metric that includes the three factors mentioned above and time. This metric normalizes each indicator and performs a weighted sum with equal weights. Specifically, for indicators like success rate, where higher values are better, the following formula is used,
\begin{equation}
    V_{norm} = \frac{V - V_{min}}{V_{Max} - V_{min}},
\end{equation}
and for indicators like collision rate, where lower values are better, the following formula is used,
\begin{equation}
    V_{norm} = \frac{V_{Max} - V}{V_{Max} - V_{min}},
\end{equation}
where $V$ represents the value of each indicators, and $V_{max}$ and $V_{min}$ are the highest and lowest ones across the approaches.
Finally, 
\begin{equation}
    ComScore = \sum_{i=1}^{n} w_i \cdot V_{norm, i}.
\end{equation}

% \subsection{Implementation Details}

% The approach is implemented with PyTorch. 
% The linear layer for condition encoding has input and output dimensions of  [4200, 60] and the hidden layer of [786, 512, 256]. We set the potential dimension to 512 and the diffusion step $T$ to 200 to train the diffusion model. We used the Adam optimizer with a learning rate of 0.0003. 
% The number of iterations is 25k. For the transformer module, we used 8 layers with a potential dimension of 512. The training batch size is set to 64. The model is executed on NVIDIA RTX A5000 GPUs.

\subsection{Results and Analysis}
\begin{figure*}[t]
    \centering
    \subfigure[Success rate]{
    \begin{minipage}[t]{0.5\linewidth}
    \centering 
    \includegraphics[width=\linewidth]{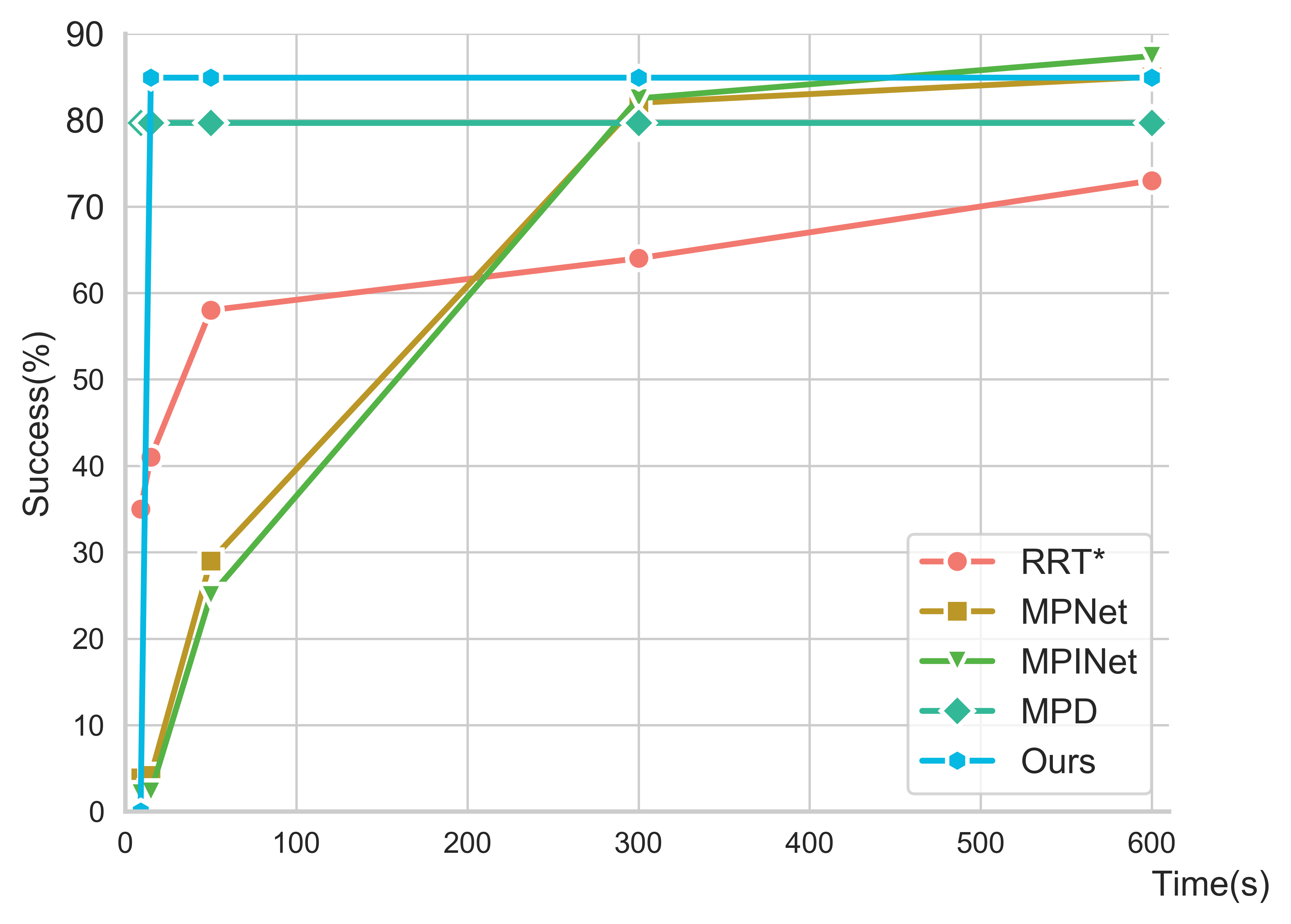}
    \label{overtime}
    \end{minipage}
    }%
    \subfigure[Visualization]{
    \begin{minipage}[t]{0.48\linewidth}
    \centering
    \includegraphics[width=\linewidth]{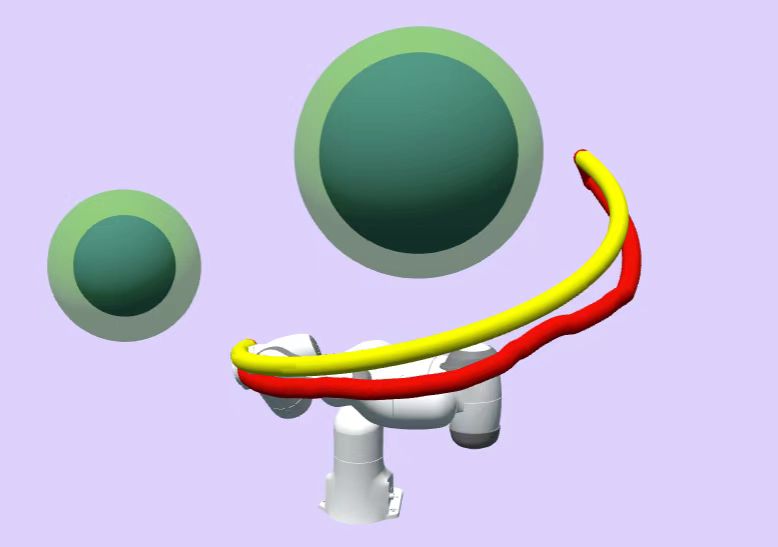}
    \label{visualization}
    \end{minipage}
    }%
    \centering
    % \vspace{-0.3em}
    \caption{Comparison of success rates over time and visualization of robot obstacle avoidance trajectories.}
    \vspace{-0.5cm}
    \label{overtime_visualization}
\end{figure*}

Table \ref{results} and Fig.~\ref{overtime} compare performance on the ROP dataset. While RRT* and MPINet achieve high success rates at 10 minutes, their latency is impractical for real-time use. In contrast, diffusion-based methods MPD and RobotDiffuse provide high-success paths in just 9–15 seconds. 
Specifically, RobotDiffuse reaches an 84.9\% success rate in 15 seconds and outperforms all baselines within 5 minutes, showing an 11.6\% improvement over traditional methods. Regarding quality, RobotDiffuse produces paths 0.819 units shorter than the best sampling-based method, demonstrating superior global optimization. Despite MPD’s slight edge in smoothness, RobotDiffuse exhibits better scenario adaptability and achieves the highest overall ComScore, offering a reliable balance of efficiency and planning.

To provide an intuitive comparison of pose generation, we illustrate the trajectory of the robot navigating through obstacles in Fig.~\ref{visualization}. 
The blue shape represents the robot's body, while the green area serves as a buffer zone to prevent collisions with obstacles. 
Compared to the red trajectory generated by MPINet, the yellow trajectory produced by our method is smoother and more efficient. The dynamic process is further demonstrated in the attached video.

\begin{table}[htbp] 
\renewcommand{\arraystretch}{1}
\centering 
\caption{Main results on ROP dataset. 
} 
\label{results}
% \vspace{5pt} 
\setlength{\tabcolsep}{1mm}{
\scalebox{1}{
\begin{threeparttable}
\begin{tabular}{l|ccc|c|c} 
\toprule 
&Suc (\%)$\uparrow^1$&Col (\%)$\downarrow$ &Len$\downarrow$ &Time (s)$\downarrow$ &ComScore$\uparrow$\\ 
\midrule
RRT* \cite{karaman2011sampling} &73.3 &\bfseries{0}  &15.037 &600 &0.4175\\
MPNet \cite{qureshi2020motion}  &84.5 &14.3 &18.492 &600 & 0.19\\
MPINet \cite{fishman2023motion} &\bfseries{87.4} &11.6 &16.438 &600 & 0.4175\\
MPD \cite{carvalho2023motion} &79.7  &12.4  &\bfseries{14.142}  & \bfseries{9} & \underline{0.645}\\
Ours &\underline{84.9} &\underline{9.8} & \underline{14.218} &15 & \textbf{0.765} \\
\bottomrule
\end{tabular}
\begin{tablenotes}
    \item[1] The arrows indicate higher/lower value corresponds to better performance.
\end{tablenotes}
\end{threeparttable}
}}
\vspace{-2em}
\end{table}

\vspace{-0.3em}

\subsection{Ablation Studies}

To systematically validate the effectiveness of our method across key components, including the point-cloud encoder, the backbone network of the diffusion model, and the introduction of physics guidance, we designed a series of ablation experiments. All experiments were trained and evaluated using the ROP dataset. The results are also shown in \fullref{Table}{abl_results}.

\begin{table}[htbp] 
\renewcommand{\arraystretch}{1}
\centering 
\caption{Ablation results.
} 
\label{abl_results}
% \vspace{5pt} 
\setlength{\tabcolsep}{1mm}{
\scalebox{1}{
\begin{threeparttable}
\begin{tabular}{l|ccc|cc} 
\toprule 
&Suc (\%)$\uparrow$&Col (\%)$\downarrow$ &Len$\downarrow$ &Time (s)$\downarrow$\\ 
\midrule
PointNet\cite{charles2017pointnet} &69.4 &26.7  &17.154 &42\\
DGCNN\cite{wang2019dynamic} &73.4 &22.5 &16.864 &64\\
\midrule
U-Net\cite{ho2020denoising} &59.3 &29.7  &17.358 &27\\
Transformer\cite{vaswani2017attention} &69.0 &16.5 &17.137 &32\\
Decoder-only &62.8 &23.8 &16.195 &23\\
\midrule
Noise &64.5 &41.6 &17.628 &15\\
Joint-only &68.1 &21.1 &16.148 &15\\
Joint\&Point &74.0 &19.5 &14.762 &15\\
JPC  &79.1 &\underline{9.2} &14.571 &15\\
\midrule
Ours &\textbf{84.9} &\textbf{9.8} &\textbf{14.218} &15\\
\bottomrule
\end{tabular}
\end{threeparttable}
}}
\vspace{-2em}
\end{table}

Ablation studies confirm the superiority of the CAE model for obstacle feature extraction over PointNet or DGCNN, which likely suffered from information loss during joint training. Regarding the diffusion backbone, the encoder-only Transformer proves most effective, leveraging its bidirectional attention mechanism to capture long-range dependencies and global motion context while streamlining computational complexity for better scalability. 
Furthermore, the integration of targeted loss components $\mathcal{L}_{\text{point}}$ and $\mathcal{L}_{\text{collision}}$ is pivotal for reducing attitude errors, while the proposed diffusion guidance significantly enhances the overall success rate by ensuring precise trajectory alignment during critical middle stages of motion, effectively balancing a marginal increase in collision frequency with a higher rate of mission completion.

\section{Conclusion}
In this work, we introduced RobotDiffuse, a diffusion-based motion planning approach for redundant manipulators that addresses the challenges of high-dimensional motion planning. By integrating physical constraints through a point cloud encoder and using an encoder-only transformer to capture temporal dependencies, RobotDiffuse enhances both the efficiency and stability of motion planning. Our approach not only overcomes the computational inefficiencies of traditional sampling-based methods but also improves the coherence and quality of the generated trajectories. The proposed method was validated using a complex simulation environment and a large-scale dataset, showing promising results in terms of feasibility and performance. These findings suggest that diffusion models hold significant potential for advancing motion planning tasks, especially in dynamic and complex environments. In addition, we have created and publicly released a trajectory optimization dataset for non-desktop scenarios. We hope this resource will support and inspire further research in the field. 
In future work, we plan to explore the application of fast diffusion methods to enhance the efficiency of motion planning. Additionally, we aim to validate our approach on real manipulators with varying numbers of joints to further assess its practicality and robustness.

\end{document}